\documentclass{article}

\usepackage[nonatbib,final]{neurips}

\usepackage[utf8]{inputenc} 
\usepackage[T1]{fontenc}    
\usepackage{hyperref}       
\usepackage{url}            
\usepackage{booktabs}       
\usepackage{amsfonts}       
\usepackage{nicefrac}       
\usepackage{microtype}      
\usepackage{xcolor}         

\usepackage{wrapfig}
\usepackage{amsmath}
\usepackage{makecell}
\usepackage{multirow}
\usepackage{witharrows}
\usepackage{annotate-equations}
\usepackage{mathtools}
\usepackage{adjustbox}
\usepackage{subcaption}
\usepackage{tabularx}
\usepackage{xcolor}

\definecolor{lightgray}{gray}{0.5}

\newenvironment{conditions}[1][where:]
  {#1 \begin{tabular}[t]{>{$}l<{$} @{${}\to{}$} l}}
  {\end{tabular}\\[\belowdisplayskip]}

\newcommand{\coloredtext}[1]{\textcolor{lightgray}{#1}}
\newcommand\Tstrut{\rule{0pt}{2.6ex}}         
\newcommand\Bstrut{\rule[-0.9ex]{0pt}{0pt}}   

\title{Sculpting Efficiency: Pruning Medical Imaging Models for On-Device Inference}

\author{
Sudarshan Sreeram \\
Imperial College London \\
{\tt\small ss8119@imperial.ac.uk} \And 
Bernhard Kainz \\
Imperial College London \\
{\tt\small b.kainz@imperial.ac.uk}
}

\begin{document}

\maketitle

\begin{abstract}
Leveraging ML advancements to augment healthcare systems can improve patient outcomes. Yet, uninformed engineering decisions in early-stage research inadvertently hinder the feasibility of such solutions for high-throughput, on-device inference, particularly in settings involving legacy hardware and multi-modal gigapixel images. Through a preliminary case study concerning segmentation in cardiology, we highlight the excess operational complexity in a suboptimally configured ML model from prior work and demonstrate that it can be sculpted away using pruning to meet deployment criteria. Our results show a compression rate of \textbf{1148x} with minimal loss in quality ($\sim 4\%$) and, at higher rates, achieve faster inference on a CPU than the GPU baseline, stressing the need to consider task complexity and architectural details when using off-the-shelf models. With this, we consider avenues for future research in streamlining workflows for clinical researchers to develop models quicker and better suited for real-world use.
\end{abstract}


\section{Introduction}

CNNs thrive in multi-disciplinary settings, but their architectural depth correlates with growing memory footprint and operational expenses \cite{menghani2021efficient}. Despite their prevalence in academic healthcare research, such models remain rare in clinical practice \cite{Kelly2019}; barring notorious regulatory hurdles, this scale-related complexity impedes effective deployment. For context, models in point-of-care (POC) systems and routine medical image assessment must maintain clinical utility while being performant in compute-limited settings. This limitation is evident as healthcare institutions seldom upgrade their IT infrastructure \cite{Zhang2022}. Moreover, domain-specific modalities yield gigapixel images \cite{Xu2017}, and operating large models may pose hardware constraints that existing IT setups cannot support; conversely, using the smallest feasible patch size may overwhelm the I/O system \cite{9835363}. Shifting tasks to cloud services can jeopardise patient data confidentiality, and investing in advanced hardware may be infeasible for underfunded hospitals, particularly in developing countries. An optimal solution permits healthcare specialists to perform intricate image analysis on various platforms, from smartphones to web browsers in hospital computers, ensuring widespread utility and data protection.

As such, we posit that this deployment inefficiency may partly stem from clinical researchers' limited time and engineering expertise in aptly configuring ML models \cite{butoi2023universeg}, often leading them to adopt a data-driven approach and favour off-the-shelf solutions \cite{Zhang2022}. The rapid pace of research in ML exacerbates this challenge. Generally, such researchers undertake proofs-of-concept with ample server resources. Nevertheless, addressing the challenge of optimising model selection and configuration can bolster stakeholder confidence, validate clinical feasibility, and even improve training efficiency.

\textit{Model compression}, specifically pruning, is a promising solution. It addresses the challenge of maintaining a model's predictive power while minimising resource usage. Pruning is an umbrella term for a broad set of methods that remove redundant network weights \cite{zhu2017prune}, and it's crucial in enabling the exciting prospect of inferencing directly on portable and tabletop instruments or even smartphones connected to medical-grade scanners like POCUS ultrasound devices. These instruments, similar to entertainment devices, capture and post-process information at high throughput with limited compute capacity for heavy ML workloads. Optimising for such devices offers reduced latency \textendash\ vital for time-sensitive emergencies \textendash\ and better reliability by eliminating the need for internet access.

\textbf{Related Works} Research on pruning models in medical imaging \cite{Agarwal2022, Rajaraman2020, 9222548, DINSDALE2022102583, sapkota2023neural} predominantly adopts a post hoc ``pruning after training'' approach. While these works have a commendable, healthcare-centric focus on preserving clinical utility to prevent misdiagnosis, they sidestep a deeper exploration into the root causes behind the excess operational complexity in models, thus missing the opportunity to critically assess and propose methodical changes to the ubiquitous, cookie-cutter approach most accessible and favoured by non-ML practitioners. The prevailing status quo is such that proactively avoiding the `\textit{mistake}' beforehand is overshadowed by subsequent correction efforts.

\textbf{Contribution} This paper dissects researchers' oversights in configuring an off-the-shelf model (DeepLabV3), revealing deep-rooted inefficiencies. While our early-stage experiments indeed adhere to a post hoc approach, we investigate and shed retrospective light on the underlying cause of these issues and consider proactive solutions to explore in a future scope. We expect such solutions to streamline workflows for clinical researchers in developing models quicker and better suited for real-world use (e.g., clinical trials and end-to-end clinical workflows).

\section{Method}


\renewcommand{\baselinestretch}{1.05}\selectfont
\textbf{Weight Pruning} Using two unstructured weight pruning techniques, we sparsify models by zeroing a subset of weights via an L1-norm saliency criterion \cite{han2015learning}. Given a sparsity budget $S \in [0, 1]$, the \textit{local} method prunes each layer's weights $W^j$ using a threshold $T^j = \text{quantile}(|W^j|, S)$, leading to $W'^j = W^j \odot \mathbf{1}(|W^j| > T^j)$. This layer-wise approach ensures the overall model's sparsity approximates $S$. Furthermore, the \textit{global} method prunes all weights $W$ against a single threshold $T = \text{quantile}(|W|, S)$, yielding $W' = W \odot \mathbf{1}(|W| > T)$, allowing layer-wise redistribution based on redundancy. Here, $\mathbf{1}(\cdot)$ represents a binary mask. While sparse tensors yield a storage benefit, computational challenges arise from the high irregularity of zeros \cite{wang2021sparsednn, mao2017exploring}. As pruned weights still exist in memory, pruned models share the same footprint (e.g., latency, MACs, size) as the baseline. We use weight pruning as a tool to reveal redundancies and disregard custom accelerators. 

\renewcommand{\baselinestretch}{1.05}\selectfont
\textbf{Filter Pruning} A more coarse-grained strategy than weight pruning, filter pruning focuses on removing entire filters from convolutional layers, leading to dense, lean networks \cite{mao2017exploring}. Mathematically, for each layer $j$, filters are pruned against a threshold $T^j = \text{quantile}(\{||W^j_i||_1\}_{i=1}^{n_j}, S)$, where $W^j_i$ represents the weights of the $i$-th filter, $S \in [0, 1]$ dictates the desired filter sparsity, and $n_j$ is the total number of filters. Post-pruning, only filters satisfying $||W^j_i||_1 > T^j$ remain in the layer. Such models structurally resemble the baseline, differing only in channel counts, and the high regularity enables the model to run on standard hardware \cite{liu2019rethinking}. We use Microsoft's NNI toolkit for this process \cite{nni}.


\section{Experiments}

\textbf{Setup} In our experiments, we adopt a one-shot pruning schedule and fine-tune the model for $5\!\sim\!10$ epochs. In each pruning run, we set the sparsity as $S = 1 - 0.5^x$, where $x$ is the run index. This formula ensures an increasing but decelerating sparsity rate; high targets can corrupt a model's predictive power. To assess model performance, especially for filter-pruned models, we run inference on both the CPU and GPU using the Intel i5 13600K (64 GB RAM) and an Nvidia RTX 4080 (16GB).

\textbf{Cardiac Ultrasound} The EchoNet-Dynamic dataset \cite{Ouyang2019EchoNetDynamicAL} comprises 10K+ cardiac ultrasound videos with a resolution of $112 \times 112$. The segmentation target is the left ventricle, a heart chamber, and the trend in its volume over time correlates with prognosis \cite{Ouyang2020VideobasedAF}. For frame-level segmentation, the dataset curators proposed a DeepLabV3-based model (39.6M params. \textendash\ 158.76 MB) with roughly 7.8 GMACs. Our baseline reproduction achieves a DICE score of $0.9098 \pm 0.0026$ and $0.9334 \pm 0.0017$ on the systolic and diastolic frames. It has a latency\,/\,throughput of 31.82 ms\,/\,31 FPS on a CPU and 5.279 ms\,/\,189 FPS on a GPU. Notably, the mean sampling rate of videos is 51 FPS, so CPU inference is unsuitable for real-time use.

\begin{figure*}[t]
\centering
\begin{center}
\begin{subfigure}[m]{0.2\textwidth}
    \includegraphics[width=\textwidth]{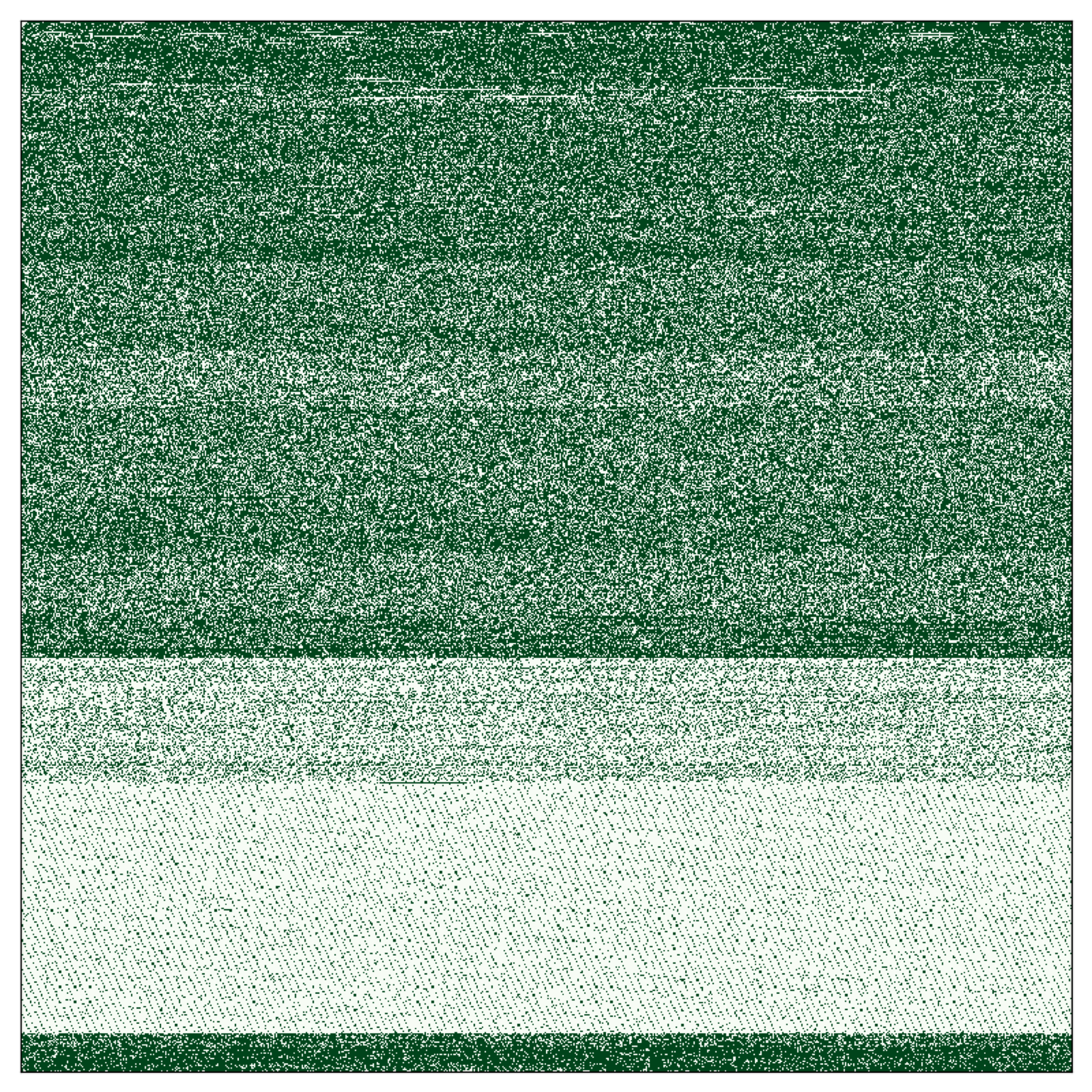}
    \vspace{0.05em}
\end{subfigure}
\hfill
\begin{subfigure}[m]{0.26\textwidth}
    \includegraphics[width=\textwidth]{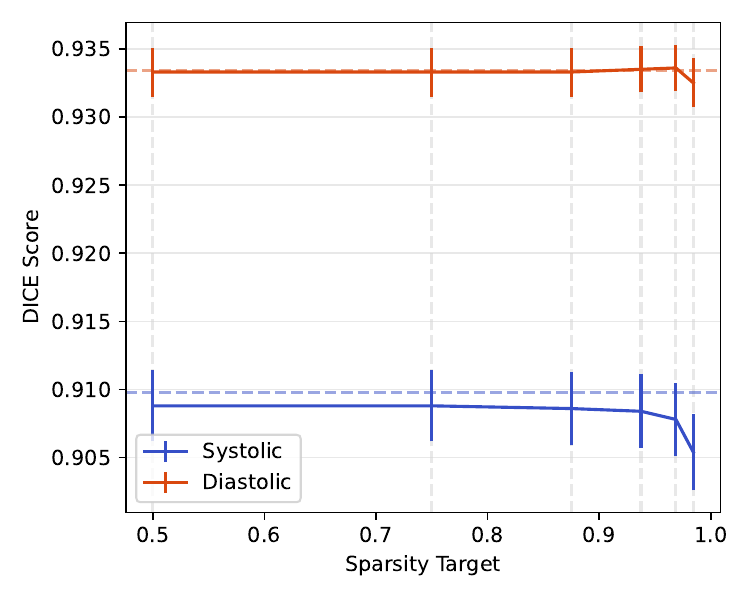}
\end{subfigure}
\hfill
\begin{subfigure}[m]{0.26\textwidth}
    \includegraphics[width=\textwidth]{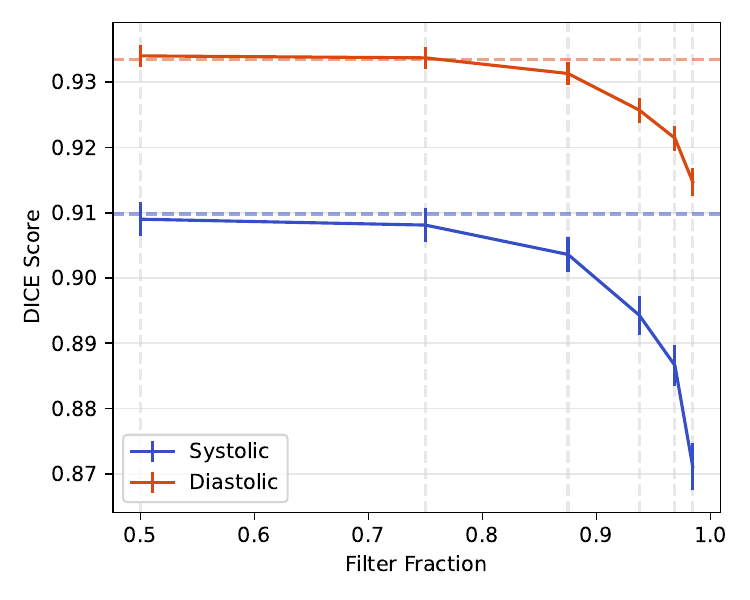}
\end{subfigure}
\hfill
\begin{subfigure}[m]{0.26\textwidth}
    \includegraphics[width=\textwidth]{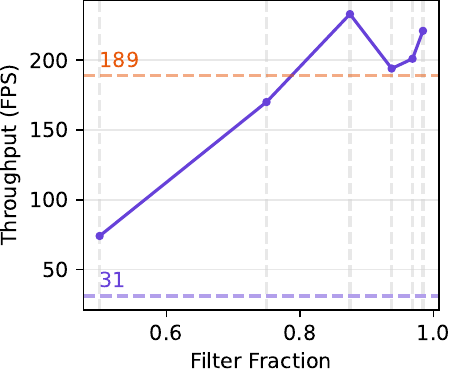}
\end{subfigure}
\end{center}
   \caption{
    Characteristics of weight and filter pruned models. \textit{Left}: Parameter sparsity for a 50\% weight pruned (global) model, with non-zero values coloured green. \textit{Middle}: DICE score trends for systolic and diastolic frames under global weight pruning (left) and filter pruning (right). \textit{Right}: Trend in CPU throughput for filter pruned models with increasing fraction of filters pruned; the dotted orange and violet markers represent the GPU and CPU baselines, respectively.
}
\label{fig:results}
\end{figure*}

In Figure \ref{fig:results}, globally weight-pruned models exhibit impeccable resiliency, maintaining quality within 1\% of the baseline even at a sparsity of 98.4\% (633K params.). On the contrary, a local strategy (not depicted) deteriorates model quality at high targets \cite{blalock2020state}; some layers are more sensitive than others. Our filter pruning results corroborate the overarching observation that DeepLabV3 is substantially overparameterized for this task. Specifically, the most extreme case, with 98.4\% filters pruned, yields DICE scores within 4\% of the baseline while being 1148x smaller (34.5K params.). Moreover, the optimal filter pruned model (0.875) deviates from the baseline by only 0.7\% while offering 7.5x higher throughput at 233 FPS, making it more than viable for real-time use. This model is just as fast on a CPU as the GPU baseline, a profound improvement that results solely from pruning.

Using an off-the-shelf model without considering task complexity is akin to using a truck to transport a grape. DeepLabV3 can, as proposed, handle 20 segmentation classes and is architected to capture multi-scale contextual information. However, the task here is a binary problem where the segmentation target is a smooth, continuous blob in largely the same position for every sample, varying little in scale as trained sonographers obtain these video samples. The parameter sparsity diagram in Figure \ref{fig:results} indicates the immediate redundancy. The missing band belongs to DeepLabV3's ASPP module, precisely three atrous convolutions (each 4.71M params.), forming 35.7\% of the model. 

Atrous convolutions excel under scale-affecting transforms (e.g., varying imaging parameters) and when the feature map size is smaller than the atrous rate. Here, the backbone encoded feature map has a resolution of $14 \times 14$, but the ASPP module is configured with atrous rates 12, 24 and 36; all three convolutions reduce to pointwise ones \cite{chen2017rethinking}. The minimum input size to circumvent this is $288\times288$, more than twice the frame size configured by the dataset curators; this also applies to their recent work, EchoNet-LVH. The oversight of not tailoring DeepLabV3 to dataset characteristics stems from the nuanced ML and engineering details underpinning this architecture. Any researcher strapped for time in delving deep into such details or navigating large, intricate codebases may have easily missed these critical requirements, as evinced in this case. 

\textbf{Conclusion} Our preliminary work uncovered latent inefficiencies tied to using DeepLabV3 off the shelf for a segmentation task in cardiology. By scrutinising and reproducing prior work, we highlighted the pivotal role of nuanced architectural details in model efficiency. Our analysis stands as a cautionary tale, and our subsequent study on segmenting the vasculature in Fundus scans, demanding intricate traces, attributes nonconformity to implicit architectural constraints and low task complexity as enabling high compression rates. We acknowledge ample prospects for future work.

Pruning at initialisation \cite{lee2019snip} and zero-cost proxies for neural architecture search (NAS) over candidate modules \cite{abdelfattah2021zerocost} are promising, proactive solutions that alleviate the need to train expensive baselines, benefiting clinical researchers with a limited resource budget, including time. We are keen to also delve into more complicated multi-class and multi-label tasks in compute-heavy domains like histopathology, providing the space to test the limits of our insights with a more diverse range of models. A broad array of promising paths await, from clinician collaborations in assessing the robustness, generalisability and utility of smaller, compressed models to lowering the barrier to entry in accessing the abovementioned solutions for non-experts.

\section{Potential Negative Impact Statement}

ML-augmented POC devices offer low-cost solutions for patients and healthcare specialists alike. They aggregate quantitative metrics to detect abnormalities, enabling early treatment measures for chronic diseases such as cardiomyopathy and diabetic retinopathy \cite{Shen2017, Ouyang2020VideobasedAF, whoCardiovascularDiseases, Steinmetz2021}. Timely detection is crucial for patient health as these conditions cause irreversible damage if left untreated. As the integration of such devices gains momentum, their accessibility underscores the need for even greater vigilance of the ML models that power them. 

For healthcare specialists, ML-based workflows enable rapid evaluation, provide precise measurements, and reduce subjectivity, thereby alleviating the burden of backlogged cases \cite{hou2023domain} and allowing them to focus on critical aspects of patient care, such as treatment planning. The intention is not to replace specialists' expertise but to accelerate and foster evidence- and analysis-backed decisions \cite{Henry2022}. Here, too, the paradigm shift necessitates discerning oversight. As the models driving these workflows advance in efficiency and practical feasibility, clinical researchers can more comprehensively evaluate them in settings mirroring their target deployment environments, thus fostering an authentic platform for rigorous validation and paving the way for fine-tuning such workflows with holistic insights on their performance and implications.  

\textbf{Robustness} Ensuring model robustness is both an ethical mandate and a critical operational necessity, given the profound implications of predictions on patient outcomes. We conducted a preliminary test of our filter-pruned EchoNet models, specifically the 87.5\% sparse weight-pruned and 87.5\% filter-pruned ones, to noise. We found filter-pruned models to excel in preserving segmentation quality for both systolic and diastolic frames despite higher noise ratios, better than both the baseline and weight-pruned counterparts. Thus, even with specialised hardware to accelerate weight-pruned models, filter-pruned ones may be inherently better. 

\textbf{Generalisability} While a model's efficacy is traditionally gauged based on performance under controlled conditions, its adaptability to the multifaceted variability of real-world medical data (e.g., different scanner types, varying patient physiology) is equally, if not more, crucial. We tested filter-pruned models from our subsequent case study on the DRIVE dataset \cite{1282003}; the models were trained on a pre-processed version of the FIVES dataset \cite{Jin2022, hou2023domain}. We found that filter-pruned models generalise better up to a point; higher filter sparsity targets resulted in poorer performance, as expected.

\bibliographystyle{unsrt}
\bibliography{neurips}

\newpage
\appendix
\begin{center}
  \Large{\textbf{Supplementary Material}}
\end{center}
\section{Background}

This section elucidates, using visual aids, the core mechanics of the two examined pruning techniques, DeepLabV3 and the functioning of atrous convolutions, including a discussion of the constraints observed when the feature map size falls below the chosen atrous rate.

\subsection{Pruning}

The inspiration behind pruning, like that of neural networks, is rooted in neurobiology. The synaptic density of various regions in our brains changes with age; during infancy, there is rapid growth, followed by synaptic pruning, analogous to a \textit{``use it or lose it''} process, towards the end of adolescence to reach a need-based optimal synapse count \cite{PeterR1979}; simply put, the notion of pruning is analogous to ridding unnecessary knowledge. Similarly, naive ML models are usually over-parameterised for the tasks they address \cite{denil2014predicting}. For a given network $f(\mathbf{X}, \mathbf{W})$, with $\mathbf{X}$ being the input and $\mathbf{W}$ the weights, we formally define pruning as a method to identify a minimal subset $\mathbf{W'}$ while maintaining model quality above a threshold such that the remaining parameters in $\mathbf{W}$ are redundant (i.e. zeroed) \cite{menghani2021efficient}. Once pruned, two metrics, compression ratio and sparsity, quantify the parameter reduction in the sparse model \cite{blalock2020state}; we formulate these metrics as follows:

\begin{minipage}{0.45\textwidth}
    \[
    \text{Compression Ratio}\ =\ \frac{|\mathbf{W}|}{|\mathbf{W'}|}
    \]
\end{minipage}
\hfill
\begin{minipage}{0.45\textwidth}
    \[
    \text{Sparsity
    }\ =\ 1 - \frac{|\mathbf{W'}|}{|\mathbf{W}|}
    \]
\end{minipage}

The defining characteristic of a pruning strategy is how it precisely computes that minimal subset. There are broadly three factors along which various such strategies differ: saliency measure, granularity, and scheduling; there are a few other factors, including parameter regrowth and rewinding, but these are out of our scope.

\begin{wrapfigure}[8]{r}{0.5\textwidth}
  \vspace*{-2em}
  \includegraphics[width=\linewidth]{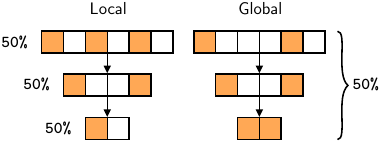}
  \caption{Comparison of pruning scopes}
  \label{fig:sparsity-scope}
\end{wrapfigure}

\textbf{Saliency Measure} Knowing or determining which parameters to prune is a fundamental decision in pruning, and a saliency measure or pruning criterion is a quantitative heuristic to evaluate the relevance of these parameters. Numerous methods exist to compute the saliency score, including magnitude-based (most common), scale-based, gradient-based and Taylor-series-based methods. Albeit simple, magnitude-based pruning is highly effective, not to mention favoured, and forms the basis for all our compression experiments. The exact formulation of this saliency measure differs with both granularity and choice of the norm (generally L1). Regardless, we invariably prune the lowest-scored weights. Furthermore, a secondary choice determines the scope of comparison: local and global. The local scope concerns ranking parameters within the same layer, while the global scope does this across the entire network; the choice of scope, in turn, determines the distribution of sparsity across the network, as visualised in Figure \ref{fig:sparsity-scope} (adapted from \cite{towardsdatascienceNeuralNetwork}).

\begin{figure}[ht]
  \centering
  \includegraphics[width=0.9\textwidth]{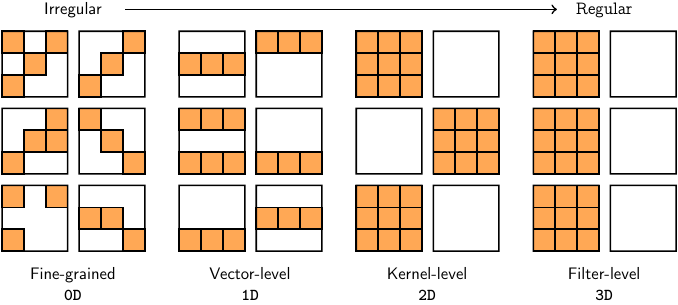}
  \caption{Granularity choices for the overall pruning strategy. Image adapted from \cite{mao2017exploring}}
  \label{fig:sparsity-granularity}
\end{figure}

\textbf{Granularity} While the saliency measure and scope address which parameters to prune and where to prune them, the choice of granularity addresses another essential question: how to prune. Predictably, multiple methods exist to offer a broad range of flexibility, as visualised in Figure \ref{fig:sparsity-granularity}. We limit our discussion to the extremes of weight and filter pruning. The former offers ample fine-grained flexibility in removing individual parameters from a network; in this particular case, the choice of a norm is irrelevant as all norms reduce to the absolute value of the parameter, preserving the ranking order. However, the same does not apply to the latter, which takes a rather aggressive, coarse-grained approach by removing entire channels; here, we instead apply the chosen norm to a group of parameters $G$, as follows:

\[
    L_1\ =\ \sum_{i \in G}|w_i| \qquad
    L_2\ =\ \sqrt{\sum_{i \in G}|w_i|^2} \qquad 
    L_p\ =\ \sqrt[\leftroot{-2}\uproot{14}p]{\sum_{i \in G}|w_i|^p}
\]

\textbf{Scheduling} Like all the other factors, scheduling also answers a few questions: when and how much to prune. Intuition from a neurobiological perspective would suggest pruning rapidly at an early stage and slowing down towards the end. While this is an excellent first guess, it is not, by any means, guaranteed to offer the best performance; a myriad of schedules exist, all with subtly different parameters. In fact, this is analogous to learning rate (LR) schedules; in both cases, one can strip away the specifics to reveal the definition of a curve that simply acts as a guide. We concern ourselves with the one-shot schedule.

Arguably, one-shot pruning is the most straightforward approach, where a trained model is pruned directly to the desired sparsity, followed by a short fine-tuning process to recover model quality; a threshold on the step count exactly determines when pruning occurs. The schedule, as a result, is a simple piecewise function formulated as follows:

\[
s_t = \begin{cases} 
      s_f &  t > t_k \\
      s_i & \text{otherwise}
   \end{cases}
\]
\begin{center}
\begin{conditions}
s_t,\ s_i,\ s_f & Current, initial and final sparsities \\
t,\ t_k & Current and threshold time steps 
\end{conditions}
\end{center}

\subsubsection{Weight Pruning}

A popular pruning strategy, supported by numerous frameworks, is unstructured low-magnitude weight pruning; the saliency measure and granularity are evident in the name. This strategy reduces the number of non-zero parameters in a network by simply setting them to zero, thus making the model weights sparse. In practice, binary masks over all parameters avoid updates during backpropagation; these masks, albeit an intuitive choice for the underlying implementation, roughly double the model size in memory. The remaining hyperparameters therein lie with the choice of scope and schedule. We formulate the simple forward pass of a linear layer with a binary mask as (adapted from \cite{julian2021stier}):
\[\underbrace{\begin{pmatrix}
0.01 & -0.4 & 1.2 \\
-1.09 & 0.35 & 0.2 \\
0.03 & 2.3 & -1.03 \\
0.7 & -0.45 & 0.82 
\end{pmatrix}}_{\mathbf{W}}
\mathbf{\odot}
\underbrace{\begin{pmatrix}
0 & 1 & 1 \\
1 & 0 & 0 \\
0 & 1 & 1 \\
1 & 1 & 1
\end{pmatrix}}_{\textbf{M}}
\cdot\  
\mathbf{x}
+ \mathbf{B}\]
\begin{center}
\begin{conditions}
\mathbf{W},\ \mathbf{M} & Weights $\in \mathbb{R}^{4 \times 3}$ and binary mask $\in \{0,1\}^{4 \times 3}$ \\
\mathbf{x} & Input $\in \mathbb{R}^{3 \times 1}$ \\
\mathbf{B} & Bias $\in \mathbb{R}^{4 \times 1}$ \\
\mathbf{\odot} & Hadamard product
\end{conditions}
\end{center}

\begin{wrapfigure}[18]{r}{0.5\textwidth}
  \vspace*{-0.5em}
  \includegraphics[width=\linewidth]{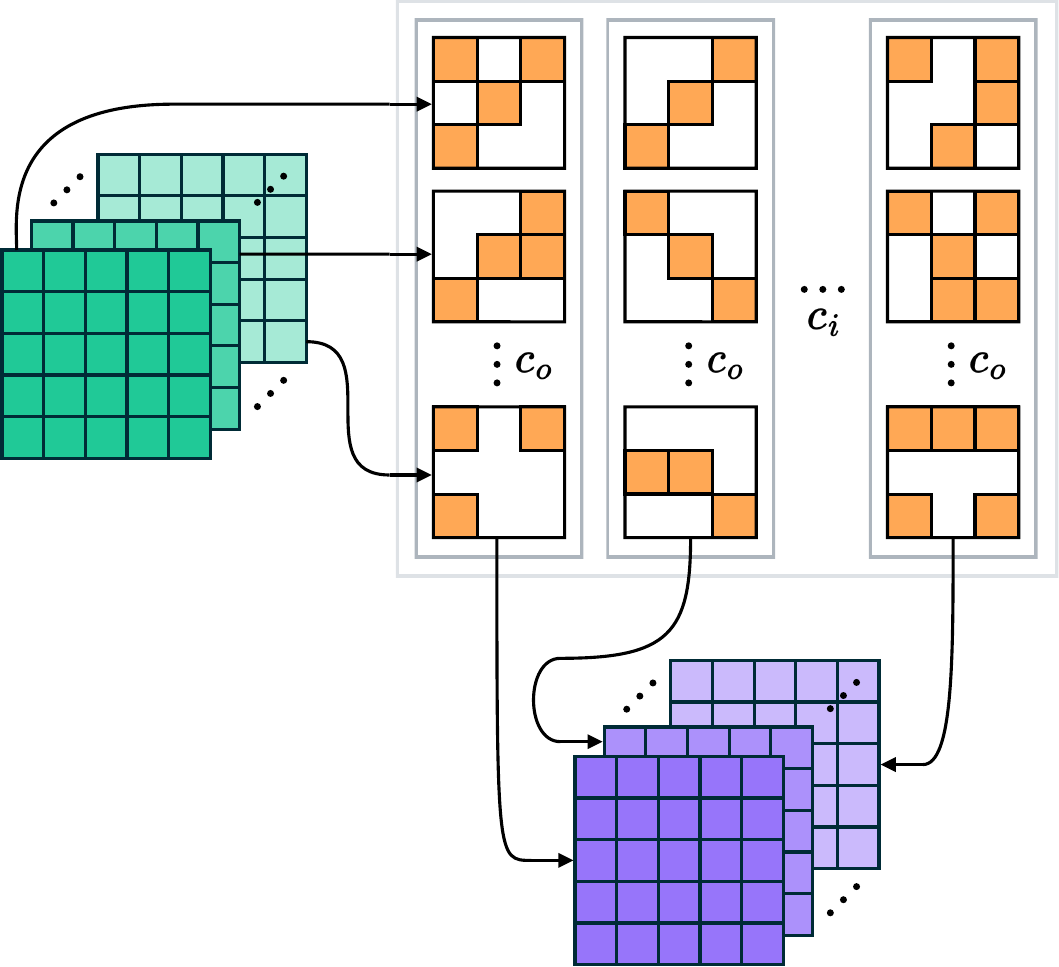}
  \caption{Unstructured weight pruning (Adapted from \cite{towardsdatascienceNeuralNetwork}); $c_i$ and $c_o$ are input and output channels. }
  \label{fig:pruning-weight}
\end{wrapfigure}

Although these sparse tensors yield a storage benefit, where, for instance, the GZIP compressed model takes up much less space, it only applies to systems with limited disk space compared to RAM, which is unlikely; note that the model is uncompressed when loaded onto working memory. Furthermore, most frameworks and general-purpose hardware cannot accelerate sparse matrices' computation \cite{wang2021sparsednn}, especially when there is no predictable, pattern-based structure; the extreme irregularity, as visualised in Figures \ref{fig:sparsity-granularity} and \ref{fig:pruning-weight}, negatively impacts the potential for trivially accelerated computation \cite{mao2017exploring}. 

In the interest of time, we deliberately avoid delving into the abyss that is the search space of pruning strategies and hardware- and software-level optimisations. We instead focus on quick, out-of-the-box methods to keep the experiments simple and easily comprehensible. With that said, we do not believe nor advocate that unstructured weight pruning is without merit, albeit in our setting, which disregards runtime optimisations, it does fall short in comparison.

\subsubsection{Filter Pruning}

\begin{figure}[h]
  \centering
  \includegraphics[width=0.9\textwidth]{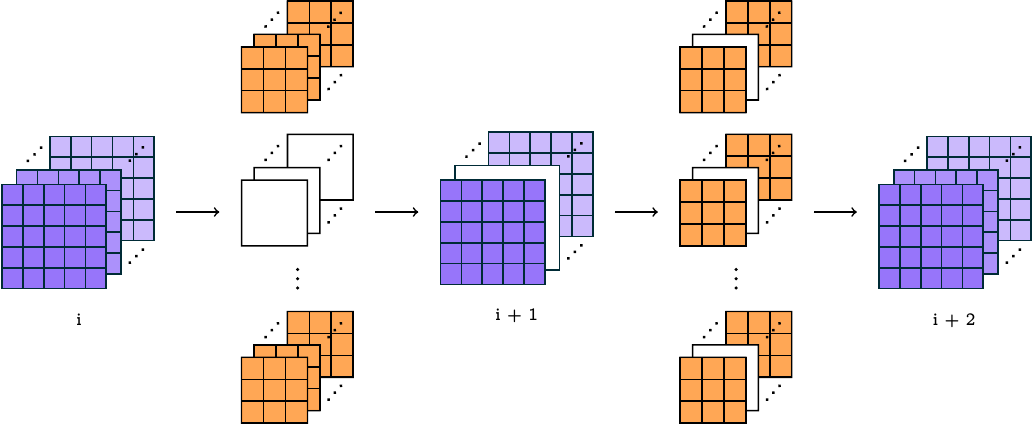}
  \caption{A visual representation of the ripple effect in structured filter pruning. A filter is pruned in layer $i$, so the corresponding feature map in layer $i + 1$ is non-existent; as a result, kernels in this layer are deleted. Image adapted from \cite{nathanhubensNeuralNetwork}.}
  \label{fig:pruning-filter}
\end{figure}

On the other end of the spectrum, structured low-magnitude filter (or channel) pruning is a highly coarse-grained strategy that removes entire filters from convolutional layers, resulting in small, dense networks \cite{mao2017exploring}; here, the notion of sparsity is non-existent in a post hoc setting as the network is structurally modified. The channel reduction in any given layer causes a ripple effect across subsequent network layers \cite{li2017pruning}, as visualised in Figure \ref{fig:pruning-filter}. Instead of independent parameters, each step prunes entire 3D sub-tensors from relevant convolutional layers, contributing to high regularity. This regularity, in turn, implies that these pruned networks neither require specialised hardware nor incur the runtime overhead of sparse inference libraries \cite{liu2019rethinking}. Notably, structurally pruned networks are architecturally similar to the baseline in their layer structure and overall flow of information, but they differ in channel counts within and across these layers.

Framework support for this pruning strategy is limited, as numerous caveats make implementation non-trivial; for instance, due to the ripple effect mentioned earlier, one would have to, say, account for channels pruned across parallel paths (e.g., projection block) to ensure compatible dimensions for subsequent operations, such as add. We used Microsoft's NNI for our pruning experiments \cite{nni}.

\subsection{DeepLabV3 \& Atrous Convolutions}

\begin{figure}[t]
  \centering
  \includegraphics[width=\textwidth]{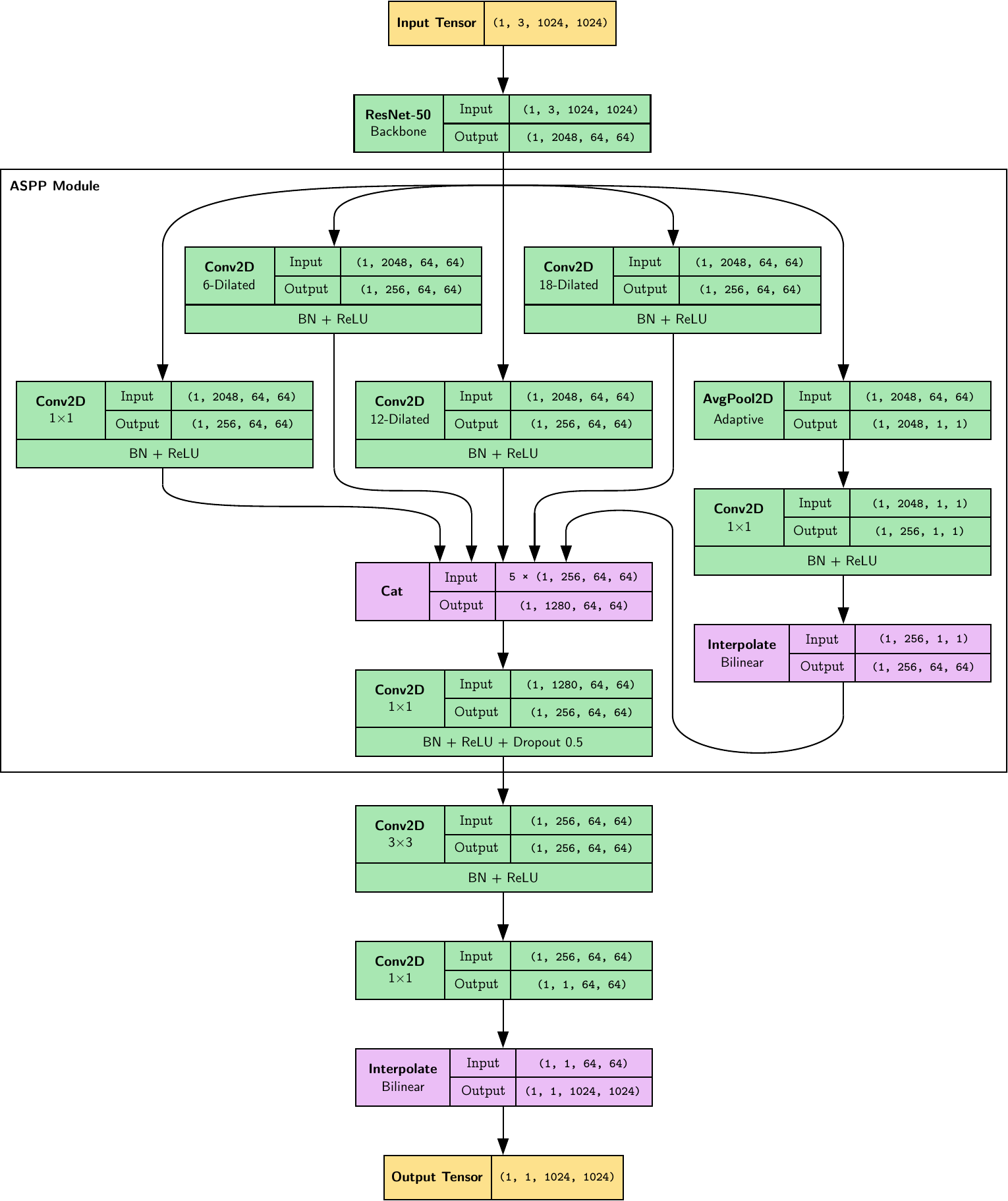}
  \caption{Architecture overview of DeepLabV3, focusing on the ``DeepLabHead''. The backbone encodes a  $1024 \times 1024$ image as 2048 $64 \times 64$ feature maps, resulting in an output stride of 16.}
  \label{fig:deeplabv3}
\end{figure}

\textbf{Output Stride} The authors of the DeepLabV3 paper noted that traditional deep convolutional networks aggressively decimate valuable, detailed information in feature maps through repeated striding or pooling operations \cite{chen2017rethinking}. To quantify this decimation, they introduce the notion of \textit{output stride}, defined as the ratio of the input spatial resolution and the final backbone-encoded feature map resolution; in Figure \ref{fig:deeplabv3}, the output stride is $1024 / 64 = 16$. Models with a low output stride tend to produce higher quality, finely detailed segmentation masks as there's more information available for reconstruction (upsampling stage). However, such models are also more resource-intensive due to the comparatively larger intermediary activations.

\begin{wrapfigure}[10]{r}{0.3\textwidth}
  \vspace*{-1.5em}
  \includegraphics[width=\linewidth]{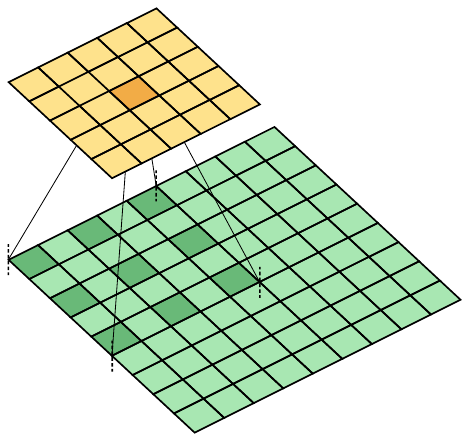}
\end{wrapfigure}

\textbf{Atrous Convolutions} Instead of using strided transposed convolutions for reconstruction (like UNets), DeepLabV3 uses an atrous spatial pyramidal pooling module \cite{chen2017deeplab}, at the core of which lies three atrous convolutions configured with different atrous rates; the convolutional layers in question form an inverted triangle in the architecture diagram visualised in Figure \ref{fig:deeplabv3}. Atrous convolutions facilitate a larger receptive field without aggressively downsampling the feature map (unlike striding) by poking holes through convolutional kernels to inflate their size. The figure on the right visualises a 2-dilated convolution on a $9 \times 9$ feature map; notice that the receptive field, demarcated by the dotted vertical markers, has a size of $5 \times5$. Varying the atrous or dilation rate allows the capture of information from a larger spatial context. The ASPP module, thus, is architected to learn multi-scale contextual information \cite{chen2017rethinking}. The choice of atrous rates impacts the module's effectiveness in capturing said information.

\begin{wrapfigure}[11]{l}{0.45\textwidth}
    \vspace*{-1.5em}
    \includegraphics[width=\linewidth]{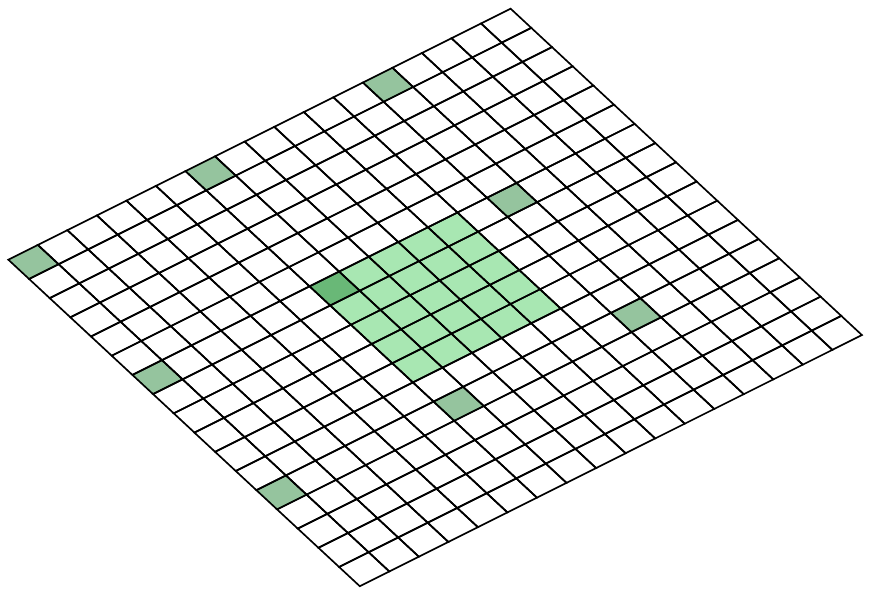}
\end{wrapfigure}

DeepLabV3, as initially proposed, uses atrous rates 6, 12, and 18 when the output stride is 16; these rates are doubled for an output stride of 8. The authors assume that with a low output stride of 8, the backbone encoded feature map resolution is high enough to use larger atrous rates 12, 24, and 36. This expectation sets a minimum constraint on the resolution of the input image: $36 \times 8 = 288$. Furthermore, they mention that it is crucial to choose atrous rates according to the output stride and feature map resolution, and there's a good reason for this: when the dilation/atrous rate grows, kernel parameters, except the central one(s), spend a growing fraction of their time \textit{``observing''} the zero-padded region rather than the feature region. In the extreme cases that this rate grows to or beyond the feature map size, the convolution degenerates to a pointwise one, as visualised in the figure on the left (6-padded, 6-dilated 3×3 conv. on a 5×5 map). The implementation details of DeepLabV3, including the choice and variant of backbone, differ, albeit slightly, across frameworks and libraries.

\section{Cardiac Ultrasound}

\begin{wrapfigure}[11]{r}{0.3\textwidth}
  \vspace*{-2.5em}
  \includegraphics[width=\linewidth]{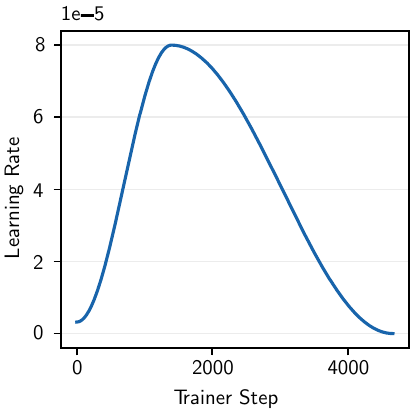}
  \caption{\texttt{OneCycle} schedule}
  \label{fig:one-cycle-lr}
\end{wrapfigure}

This section details finer details of our experiments in the first study, including some notes about our reproducing the EchoNet authors' results. 

\textbf{Training \& Baseline} Considering the lack of well-documented hyperparameters for the final EchoNet segmentation model, we started with mere default values from their original codebase. Consequently, we took the opportunity to experiment with efficient training techniques, including the use of cyclical learning rate schedules (e.g., \texttt{OneCycleLR} and \texttt{CosineAnnealingWarmRestarts}). The \texttt{OneCycleLR} scheduler contributed to efficient training and rapid iteration to quickly explore a constrained search space of hyperparameter choices; one only needs a fraction of the typical epoch count to achieve the same quality (i.e. 10 vs. 50 or even 100 epochs). This scheduler gradually increases the learning rate for a few epochs (30\% of the total pre-defined count, by default), slowly degrading the rate to zero for the remainder of the training process; the notion behind this schedule is that a high momentum earlier during training can help the model overcome local minima. For instance, Figure \ref{fig:one-cycle-lr} details a visualisation of the curve that this learning rate produces.

On a high level, our baseline uses a training budget of 10 epochs, configures the model with an SGD optimiser and \texttt{OneCycleLR} scheduler ($8 \times 10^{-5}$ peak learning rate), and uses a batch size of 16 for the data module. Table \ref{table:baseline-performance} documents the quality metrics of our baseline model, which improves on the DICE scores relative to the original baseline from the EchoNet authors. On the other hand, Table \ref{tab:inference-metrics} documents the performance metrics (i.e. inference latency and throughput on both CPU and GPU), and Table \ref{tab:baseline-misc} details other miscellaneous information.

\begin{table}[t]
\caption[justification=left]{Baseline model quality on the test set for both systolic and diastolic frames; the light grey number in the brackets represents the error margin for the 95\% confidence interval.}
\label{table:baseline-performance}
\centering
\medskip
\begin{tabularx}{\textwidth}{@{}l>{\centering\arraybackslash}X>{\centering\arraybackslash}X>{\centering\arraybackslash}X>{\centering\arraybackslash}X>{\centering\arraybackslash}X>{\centering\arraybackslash}X@{}}
\toprule
\textbf{Frame} & \textbf{AHD} & \textbf{AUC} & \textbf{IoU} & \textbf{DICE} & \textbf{Specificity} & \textbf{Sensitivity} \\ \midrule
Systolic & \makecell{$3.9619$ \\ \coloredtext{$(\pm\,0.1221)$}} & \makecell{$0.9595$ \\ \coloredtext{$(\pm\,0.0016)$}} & \makecell{$0.8435$ \\ \coloredtext{$(\pm\,0.0041)$}} & \makecell{$\mathbf{0.9098}$ \\ \coloredtext{$(\pm\,0.0026)$}} & \makecell{$0.9925$ \\ \coloredtext{$(\pm\,0.0004)$}} & \makecell{$0.9242$ \\ \coloredtext{$(\pm\,0.0035)$}} \\ \addlinespace[0.5ex]
Diastolic & \makecell{$3.8758$ \\ \coloredtext{$(\pm\,0.1002)$}} & \makecell{$0.9596$ \\ \coloredtext{$(\pm\,0.0015)$}} & \makecell{$0.8785$ \\ \coloredtext{$(\pm\,0.0027)$}} & \makecell{$\mathbf{0.9334}$ \\ \coloredtext{$(\pm\,0.0017)$}} & \makecell{$0.9929$ \\ \coloredtext{$(\pm\,0.0004)$}} & \makecell{$0.9248$ \\ \coloredtext{$(\pm\,0.0029)$}} \\ \bottomrule
\end{tabularx}
\end{table}

\begin{wraptable}[16]{r}{0.5\textwidth}
\vspace*{-1em}
\caption{Inference latency and throughput of original and scripted models on CPU and GPU.}
\label{tab:inference-metrics}
\centering
\begin{tabularx}{\linewidth}{@{}l@{\hspace{5pt}}>{\centering\arraybackslash}X@{\hspace{5pt}}>{\centering\arraybackslash}X@{\hspace{5pt}}>{\centering\arraybackslash}X@{}}
\toprule
\textbf{Model}            & \textbf{Platform} & \begin{tabular}[c]{@{}c@{}}\textbf{Latency}\Tstrut\\ (ms)\Bstrut\end{tabular}           & \begin{tabular}[c]{@{}c@{}}\textbf{Throughput}\Tstrut\\ (FPS)\Bstrut\end{tabular} \\ \midrule
\multirow{2}{*}[-0.75em]{Original} & CPU               & \makecell{$39.769$\\ \coloredtext{$(\pm 0.127)$}} & 25                        \\
                          & GPU               & \makecell{$5.373$\\ \coloredtext{$( \pm 0.002)$}}  & 186                       \\ \cmidrule(l){2-4} 
\multirow{2}{*}[-0.75em]{Scripted} & CPU               & \makecell{$31.822$\\ \coloredtext{$( \pm 0.099)$}} & 31                        \\
                          & GPU               & \makecell{$5.279$\\ \coloredtext{$( \pm 0.0)$}} & 189                       \\ \bottomrule
\end{tabularx}
\end{wraptable}

Our inference routine gathers performance metrics for both a model and its scripted counterpart under different context modes; we create the scripted model using TorchScript, an intermediate representation used for high-performance deployment, and wrap it with a \texttt{torch.jit.optimize\_for\_inference} call, which includes device-specific optimisations (e.g., operator fusion). The default mode (i.e. no context), used for training models, stores all intermediary activations and miscellaneous buffers, while the \texttt{torch.no\_grad} and \texttt{torch. inference\_mode} contexts forgo these in favour of a lightweight inference pipeline. While we measure both latency and throughput across all three, we only report metrics observed under the \texttt{torch.inference\_mode} context. Notably, the EchoNet authors do report the inference latency and runtime memory usage of their models, but we believe that both are incorrectly measured. The former uses \texttt{time.time()} for measuring the latency of GPU inference, and the latter uses Torch CUDA's memory allocation statistics, which is reportedly inconsistent. Nevertheless, they report a latency of 14 ms for an inference pass on an Nvidia 1080 Ti GPU. However, they do not detail how many samples they batch; for instance, we perform frame-wise inference (i.e. unit batch size). 

\begin{wraptable}[9]{r}{0.4\textwidth}
\vspace*{-1em}
\caption{Miscellaneous information}
\label{tab:baseline-misc}
\centering
\begin{tabular}{@{}ll@{}}
\toprule
Sparsity           & 0.0         \\
Parameters (Total) & 39,633,729  \\
Parameters (NNZ)   & 39,633,728  \\
Buffers            & 56,764      \\
MACs               & $7.827 \times 10^9$ \\
Model Size (MB)    & 158.762  \\ \bottomrule
\end{tabular}
\end{wraptable}

Considering that the frame rate of the videos is 51 frames per second, CPU-only inference for the baseline model is unsuitable for real-time usage, at least in a case without additional optimisations, including quantisation; on the other hand, the throughput on the GPU suggests that this platform option is more than capable of real-time inference, but it is worth acknowledging that such hardware may not be accessible in clinics. The common issue with running CPU-only inference on such sizeable models is that the overhead of memory accesses usually dominates latency; DRAM access is roughly 128 times slower than SRAM cache, and the associated energy usage is equally higher \cite{han2015learning}. 

\textbf{Weight Pruning} Beyond the concise experimental details and results presented in the main text, this section delves into aspects omitted in the main body due to space limitations. We ran two sets of low-magnitude pruning experiments: local one-shot and global one-shot. In all our runs, we prune and fine-tune for five epochs; that is, at the end of the first epoch, the model undergoes pruning, after which it fine-tunes for the remaining four epochs to recover quality. Furthermore, we use an \texttt{ExponentialLR} learning rate scheduler, which gradually decays the learning rate every epoch with a decay of 0.5; this scheduler has shown to be quite valuable for fine-tuning \cite{zhu2017prune}. 

\begin{figure}[t]
\vspace{-0.5em}
\centering
\begin{subfigure}[b]{0.45\linewidth}
  \captionsetup{justification=centering}
  \includegraphics[width=\textwidth]{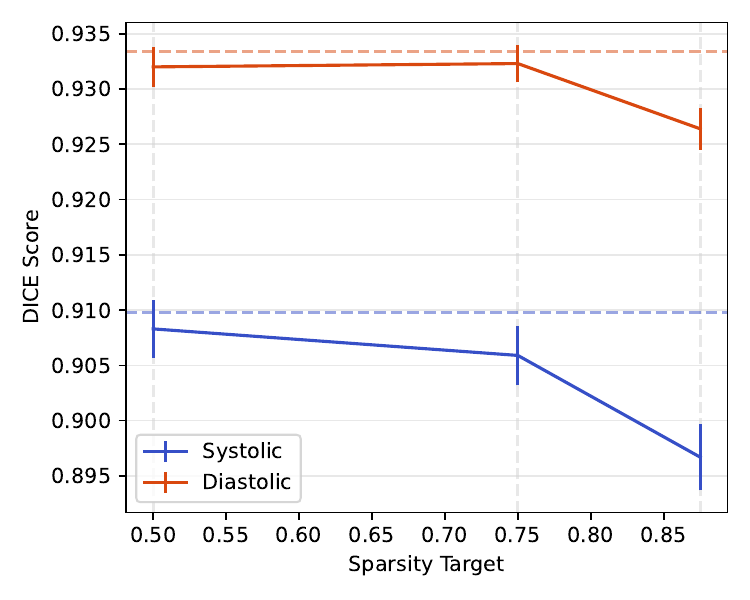}
  \caption{Local one-shot pruning at 50\%, 75\% and 87.5\% sparsity targets}
  \label{fig:local-one-shot}
\end{subfigure}
\hfill
\begin{subfigure}[b]{0.45\linewidth}
    \captionsetup{justification=centering}
    \includegraphics[width=\textwidth]{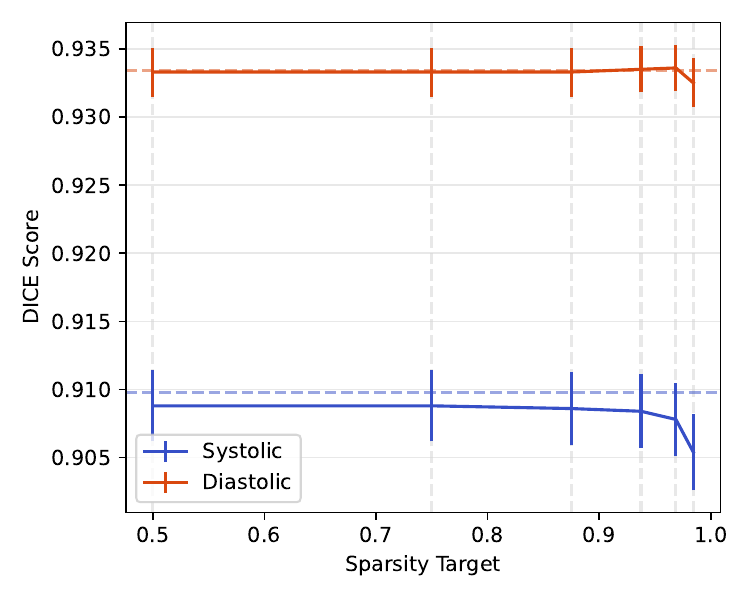}
    \caption{Global one-shot pruning at 50\%, 75\%, 87.5\%, 93.7\%, 96.8\%, and 98.4\% sparsity targets}
    \label{fig:global-one-shot}
\end{subfigure}
\caption{DICE score on the diastolic and systolic frames vs. sparsity targets for two groups of weight pruning experiments. The dotted lines represent the baseline performance.}
\label{fig:weight-pruning-results}
\end{figure}

\begin{wrapfigure}[14]{r}{0.45\textwidth}
  \includegraphics[width=\linewidth]{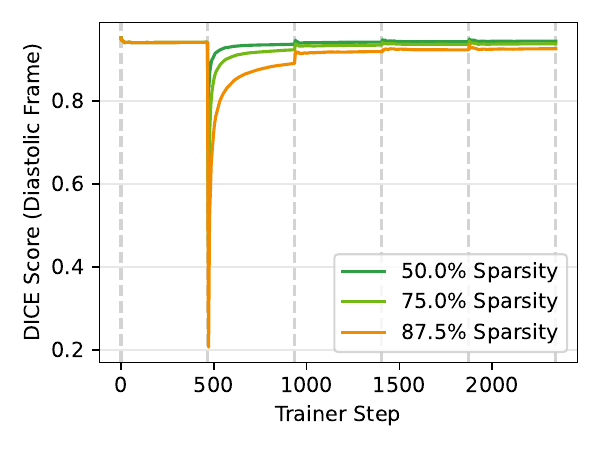}
  \vspace*{-2em}
  \caption{Recovery trend from catastrophic degradation in locally-pruned models}
  \label{fig:recovery-plot}
\end{wrapfigure}

Figure \ref{fig:weight-pruning-results} details the results from our experiments, visualising the trend in the DICE score for both frame types against sparsity; these plots detail the degradation (or improvement) in quality as the model grows sparse. As mentioned earlier, uniformly pruning all network layers is more detrimental to model quality than globally distributing the budget \cite{blalock2020state}. The step-level training plot in Figure \ref{fig:recovery-plot}, which reveals that locally-pruned models must recover from severe quality degradation, backs up this observation. Furthermore, in the same figure, the DICE score of the case with 87.5\% sparsity dips to 0.2; with a sparsity of 93.75\% (not depicted), this score hits 0, and subsequent fine-tuning fails to recover the model quality.  

\begin{figure}[h]
  \centering
  \includegraphics[width=0.815\textwidth]{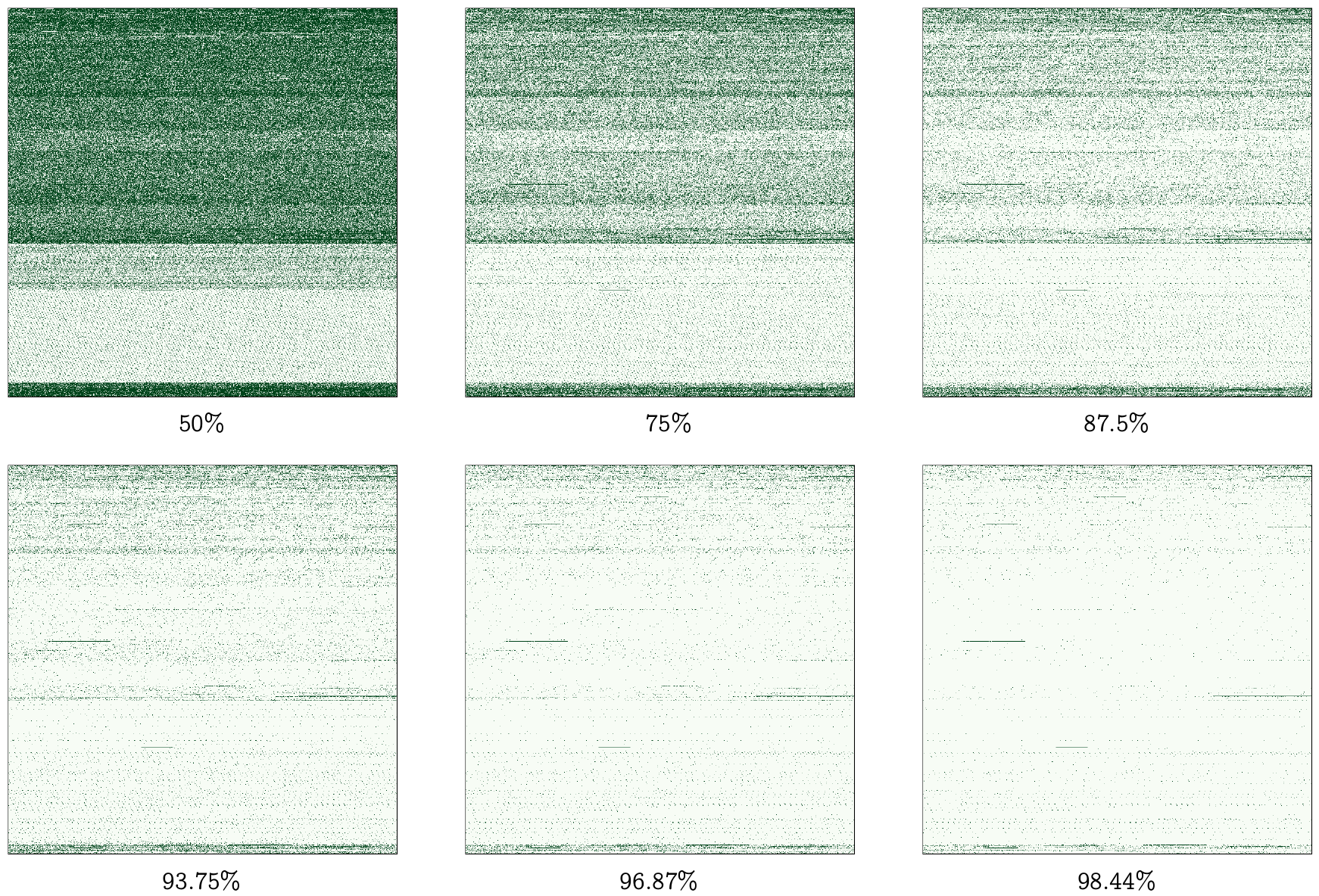}
  \caption{Parameter sparsities for globally one-shot pruned models; under each diagram is the target sparsity of the model. We do not prune biases as they are quite sensitive; the long continuous lines, particularly apparent in the case with 98.4\% sparsity, correspond to the bias parameters.}
  \label{fig:global-pruning-sparsities}
\end{figure}

\begin{wrapfigure}[31]{r}{0.5\textwidth}
    \centering
    \vspace*{-2em}
    \includegraphics[width=\linewidth]{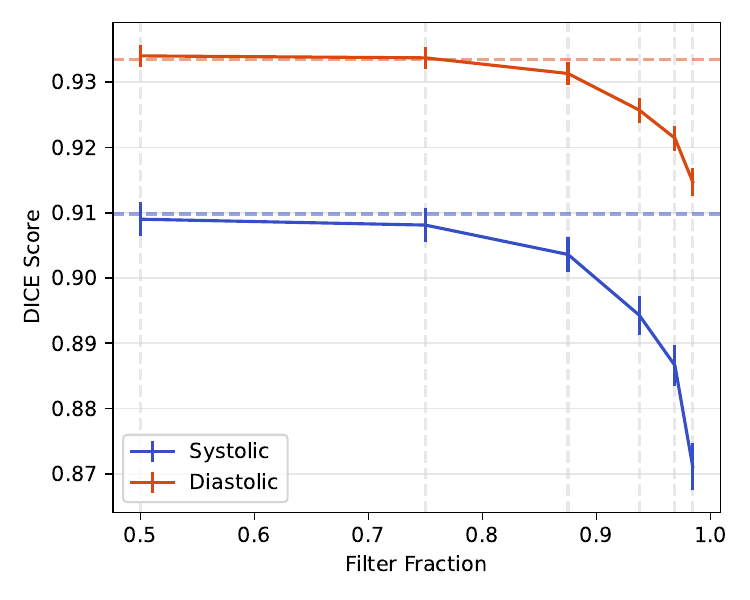}
    \caption{Trend in DICE scores of systolic and diastolic frames vs. increased compression rate}
    \label{fig:filter-pruned-perf}
    \includegraphics[width=\linewidth]{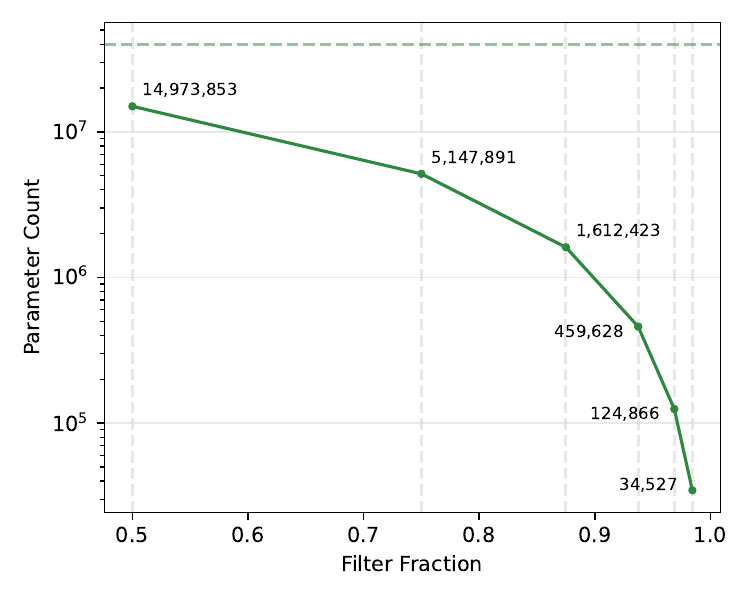}
    \caption{Parameter count vs. compression rate}
    \label{fig:param-count}
\end{wrapfigure}

\textbf{Filter Pruning} We conducted six runs in the structured pruning setting. Microsoft NNI's pruning routine does a pass over the model and constructs binary masks, which a subsequent speed-up routine uses to modify the network structure; this speed-up routine keeps track of and adjusts the input and output dimensions of other layers (e.g., \texttt{ReLU} and \texttt{BatchNorm2d}). Furthermore, we use a \texttt{OneCycleLR} schedule for the fine-tuning process with a peak threshold of $8 \times 10^{-5}$; however, we lower the peak threshold by $1 \times 10^{-5}$ with every run as we noticed that runs with a higher compression rate are more sensitive to higher learning rates. 

Figure \ref{fig:filter-pruned-perf} details the trend in quality with a higher compression rate. Notably, we use the term ``filter fraction'' instead of ``sparsity target'', as with filter pruning, we choose to remove a pre-determined fraction of filters across the entire network; it is a simple change of name, and the underlying formulation remains intact. As mentioned in the main text, the extreme case, where 98.4\% of filters are pruned, drops in quality by 4\% points relative to the baseline. We are not qualified to gauge whether this arbitrary threshold is reasonable in quantifying acceptability in a clinical environment; however, considering the integrity of the segmentation maps, we \textit{assume} that acceptance of quality seems plausible. 

\begin{figure}[b]
\centering
\vspace*{-1.75em}
\begin{subfigure}[b]{0.49\linewidth}
  \captionsetup{justification=centering}
  \includegraphics[width=\textwidth]{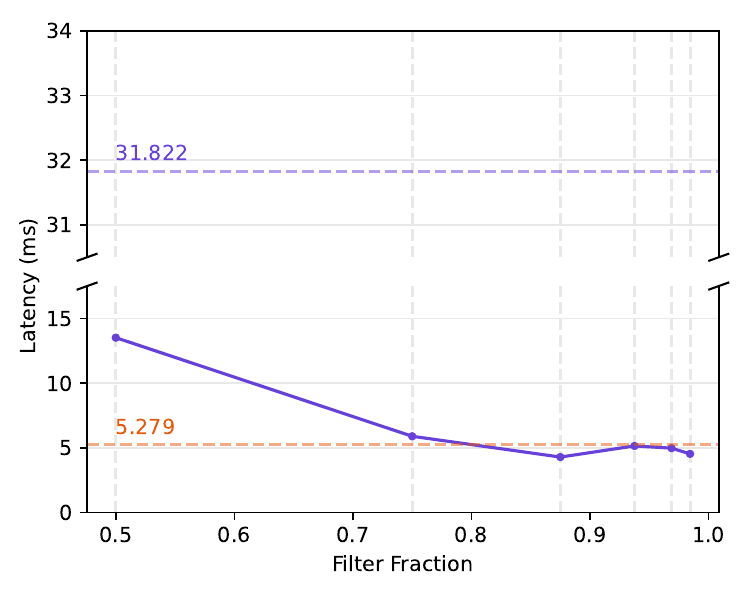}
  \caption{Trend in inference latency vs. compression rate}
  \label{fig:inference-cpu}
\end{subfigure}
\hfill
\begin{subfigure}[b]{0.49\linewidth}
    \captionsetup{justification=centering}
    \includegraphics[width=\textwidth]{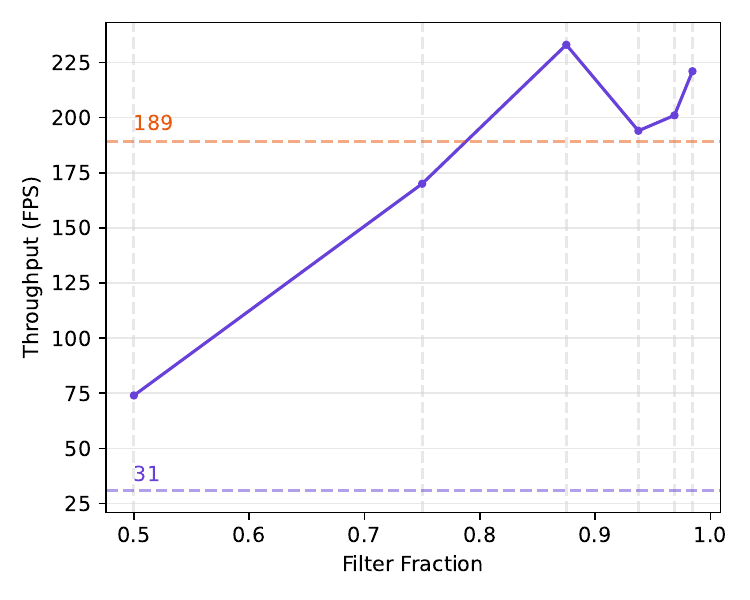}
    \caption{Trend in throughput vs. compression rate}
    \label{fig:throughput-cpu}
\end{subfigure}
\caption{Trends in the CPU-only inference latency and throughput for all six experiment runs. The dotted orange and violet markers represent the GPU and CPU baselines, respectively. For the inference latency, the error margins were too insignificant to visualise on a plot.}
\label{fig:latency-throughput}
\end{figure}

Figure \ref{fig:param-count} visualises the trend in the network's parameter count with each run. Notice that the case with a filter fraction of 0.984 only has 34.5K parameters, translating to a compression rate of \textbf{1,148}. Figure \ref{fig:latency-throughput} details the trends in latency and throughput for CPU-only inference; the GPU-only ones are monotonic, not warranting a compelling discussion. In both plots, a rather intriguing, unexpected shift occurs where, beyond a filter fraction of $0.875$, performance metrics drop.

\begin{wrapfigure}[28]{r}{0.4\textwidth}
  \centering
  \includegraphics[width=\linewidth]{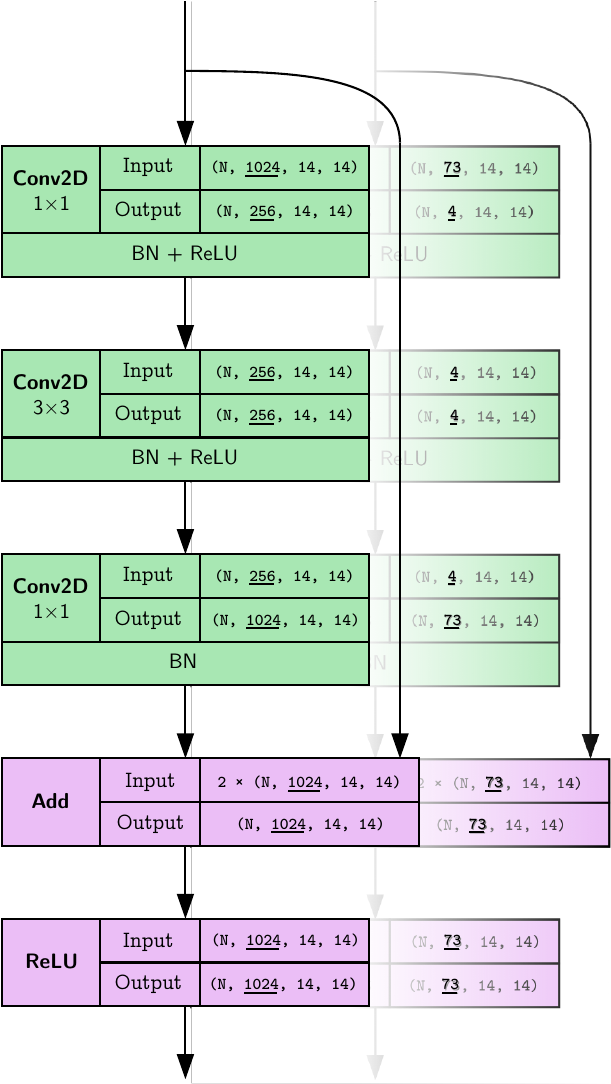}
  \caption{Bottleneck block for the baseline (foreground) and structurally pruned ($\approx 0.984$) models}
  \label{fig:structured-pruning-difference}
\end{wrapfigure}

We presume this may be because the workload is too tiny that it cannot leverage parallelism across all CPU cores. It could also be due to inefficient memory access patterns or that inference optimisations mainly focus on much larger models, leaving performance at the table for smaller ones. Identifying the exact reason warrants further study, but these plausible reasons serve as a starting point.

\begin{figure}[t]
\vspace*{-1em}
\centering
\includegraphics[width=0.9\textwidth]{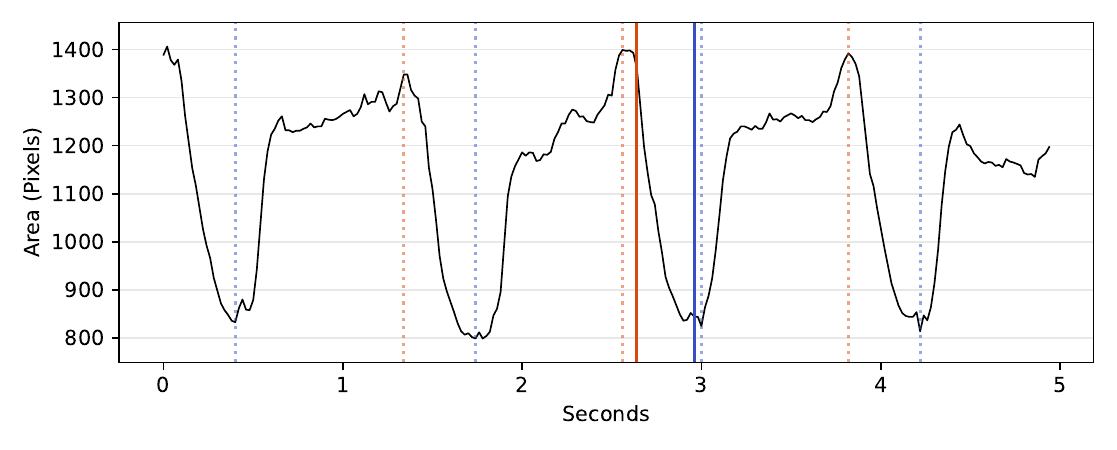}
\vspace*{-1.4em}
\caption{Post-processed frame-wise segmentations to visualise trends in ventricle area}
\label{fig:pruned-inference-area}
\includegraphics[width=0.9\textwidth]{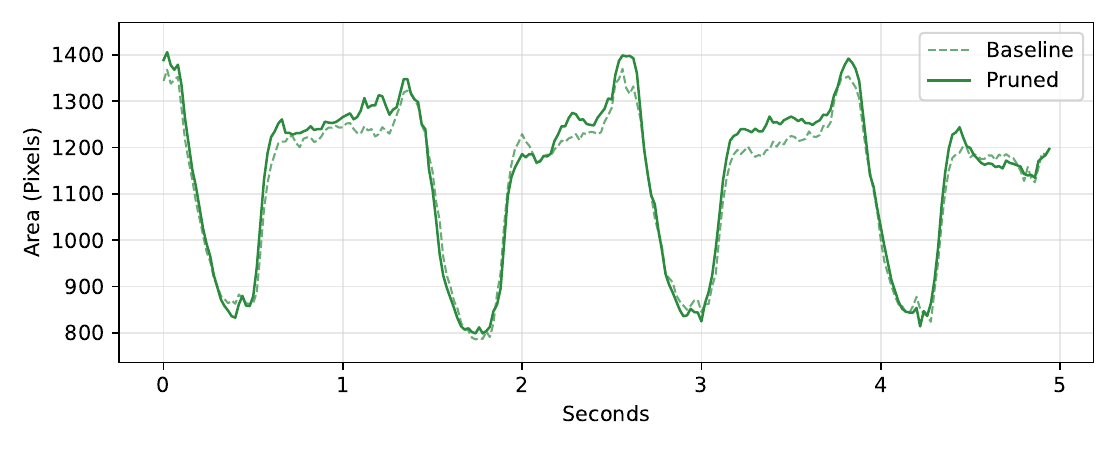}
\vspace*{-1.4em}
\caption{Deviation in ventricle area trends for baseline and pruned (0.875) models}
\label{fig:baseline-v-pruned}
\end{figure}

Figure \ref{fig:structured-pruning-difference} details a fundamental observation that filter-pruned models have the same layer structure as the baseline and only differ in that they are leaner. Figures \ref{fig:pruned-inference-area} and \ref{fig:baseline-v-pruned} showcase the trend in the ventricle area, computed through the filter pruned model's (0.875) predictions, across all frames of a sample echocardiogram video. The former showcases systolic and diastolic markers computed by post-processing the mask areas to find peaks and valleys, while the latter represents the deviation between the baseline and pruned models. We see this as another way to measure a model's predictive power qualitatively.

\textbf{Discussion} In our investigation to understand the researchers' oversight in configuring the models, we consider two factors: priorities and documentation. The EchoNet authors concern themselves with curating a dataset and evaluating the augmented clinical workflow, including studies into inter-observer variability and generalisability to samples from different hospitals. So, a standard, off-the-shelf segmentation model may have been appropriate for their use case and training budget. Nevertheless, this only partially explains the apparent oversight; this is where the second factor steps in. PyTorch's Torchvision model hub and accompanying documentation make no mention of both the underlying output stride assumption. 

\begin{wrapfigure}[19]{r}{0.4\textwidth}
  \centering
  \vspace*{-1.5em}
  \includegraphics[width=\linewidth]{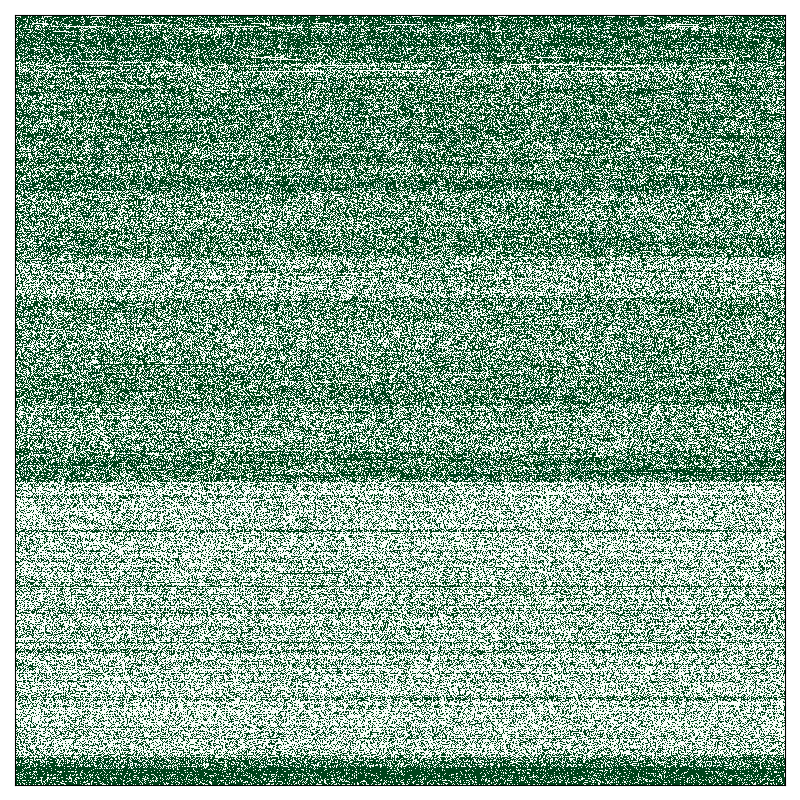}
  \caption{Parameter sparsity diagram for 50\% weight pruned DeepLabV3 with modified atrous rates.}
  \label{fig:modified-sparsity}
\end{wrapfigure}

After scouring the Torchvision codebase for answers on the implementation specifics of the ASPP module, we identified that they configure DeepLabV3 with \textit{hard-coded} atrous rates 12, 24, and 36. Consequently, the ResNet-50 backbone was configured with an output stride of 8, producing a feature map of size $14 \times 14$, smaller than the choice of rates. Again, any researcher who neither has the time to dive deep into architectural details nor navigate intricate source code may have easily missed these details.

We conducted a preliminary experiment to understand how results would change by modifying these atrous rates manually to significantly lower values: 2, 4, and 6. Although there was not much difference in performance, presumably due to the aforementioned lack of variety in scale across samples, the sparsity pattern across the band from earlier is much denser, as seen in Figure \ref{fig:modified-sparsity}; specifically, the three atrous convolutions are 67.3\%, 63.4\% and 66.9\% sparse compared to 78\% from earlier. 

\section{Fundus Retina Scans}

\begin{wrapfigure}[29]{r}{0.4\textwidth}
\vspace*{-4em}
\centering
    \includegraphics[width=\linewidth]{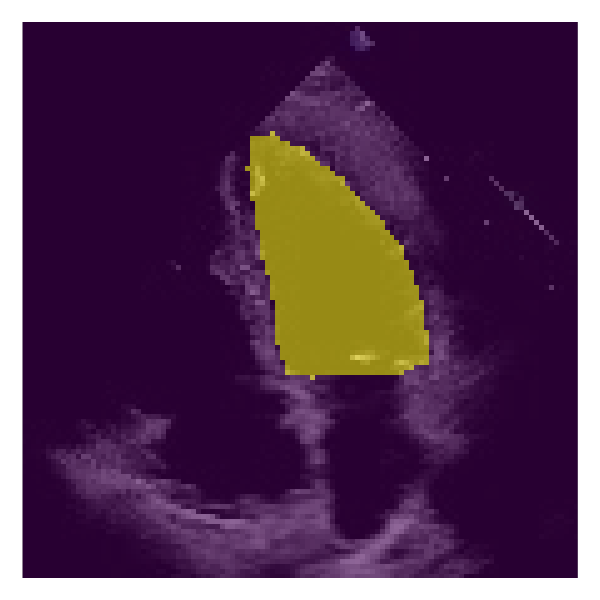}
    \includegraphics[width=\linewidth]{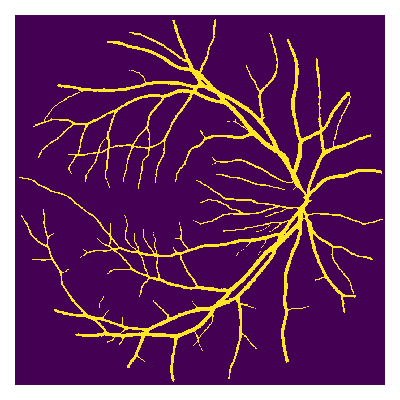}
   \caption{Sample diastolic frame segmentation (\textit{left}) vs retinal vasculature segmentation (\textit{right}).}
\label{fig:echo-segmentation}
\end{wrapfigure}

A notable follow-up question to address from our initial case study is whether a filter pruning approach, such as the one presented, would scale well to other datasets and architectures. An exciting observation prevailed as we pondered this question. The EchoNet segmentation task is fundamentally simple, as the target segmentation is a single, smooth and continuous blob in roughly the same position for every sample, as visualised in Figure \ref{fig:echo-segmentation}. We use a pre-processed version of the FIVES dataset formed of fundus photographs with a resolution of $1024\times1024$ \cite{Jin2022, hou2023domain}, larger than the minimum for DeepLabV3. The encoded feature map size is $128\times128$. While segmenting the vasculature is still a binary problem and continuous, the vessels vary in shape and size. As images of the left and right eye are flipped, the model must adapt to variations in the orientation of the overall vasculature, arguably making the task much harder than before. 

\textbf{Baseline} Our approach largely remains the same with training the baseline; we train for ten epochs using the same SGD and OneCycleLR schedule as before, albeit configured with slightly different parameters. Each sample, now 83.55x larger, occupies 12 MB, resulting in 654.25 GMACs due to sizeable activations, 93x higher than before. Since we accumulate pixel-level losses, we adjusted the learning rate to be correspondingly lower. Furthermore, this large image size limited our batch size for training; initially, we could only afford a batch size of two before running out of memory. Thus, we used 16-bit mixed precision training, which pushed our allowance to four.

Our baseline achieves a DICE score of $0.8582\pm0.0118$. The model has a latency / throughput of 2329.625 ms / 0 FPS on a CPU and 60.49 ms / 17 FPS on a GPU. Furthermore, the MAC count is now $654.253 \times 10^9$. Fundus cameras are vital in combating the cause of preventable blindness, especially in underserved regions, so efficiency is crucial, even if real-time use is optional. For instance, in contrast to clunky table-top hardware, smartphone-based ophthalmic cameras use a smartphone with a custom lens adaptor to image the retina with acceptable quality \cite{Panwar2016, Shen2017}. Similar solutions for echocardiograms exist with portable hand-held scanners that interface with smartphones. These tools are invaluable in contributing to an efficient diagnostic workflow in remote settings. 

\begin{figure}[h]
\begin{center}
\includegraphics[width=\linewidth]{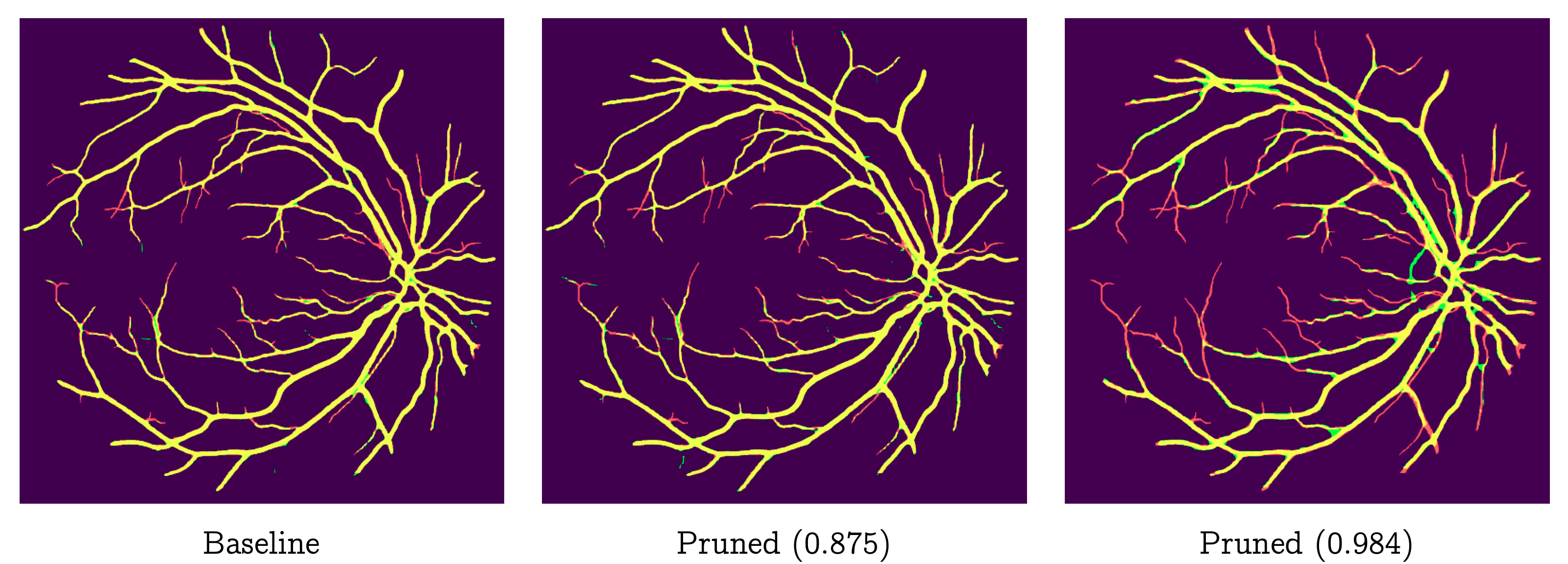}
\end{center}
   \caption{
   Difference images with respect to the ground truth for segmentations from the baseline and pruned models. \textit{Red}: Missing in predictions but present in ground truth. \textit{Green}: Present in predictions but missing in ground truth.
}
\label{fig:difference-images}
\end{figure}

\textbf{Filter Pruning} The quality trends are largely similar, albeit with a larger drop at high filter fractions. The best case (0.875) is 22.5x smaller (1.75M params.) and within 3\% of the baseline. Unlike before, the CPU performance is monotonic, with this model having a latency / throughput of 234.73 ms / 4 FPS, nearly 10x faster than the baseline. Figure \ref{fig:difference-images} reveals the integrity of its prediction with a noteworthy observation: fine details vanish as more filters are removed. This is clearly visible in the 0.984 case, achieving 66 FPS on a CPU, beating the GPU baseline. A practitioner can overcome the quality trade-off with post-processing (e.g., extracting key point descriptors). While we meet the necessary conditions, the ASPP module still remains largely redundant. The excess capacity for accommodating 20 classes is likely a dominant factor, but we defer further exploration. Extending the baseline training to reach higher quality may also provide a better starting point for pruning.

\section{AI Safety}

This section covers a few additional details surrounding our preliminary experiments examining the robustness and generalisability of the filter-pruned models from both of our studies. 

\begin{figure}[b]
  \centering
  \includegraphics[width=\textwidth]{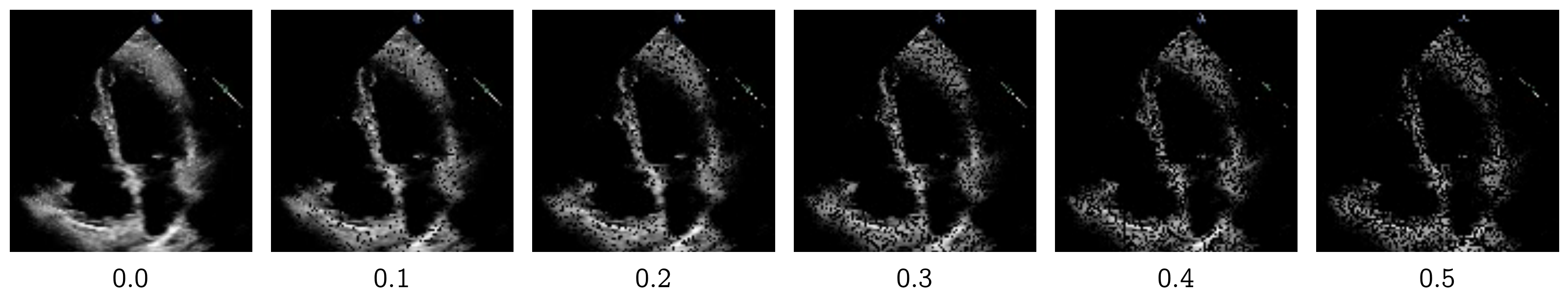}
  \caption{Visual degradation in an echocardiogram image with increasing noise}
  \label{fig:echo-noise-samples}
\end{figure}

\begin{figure}[t]
    \centering
    \begin{minipage}{0.495\textwidth}
        \centering
        \includegraphics[width=\linewidth]{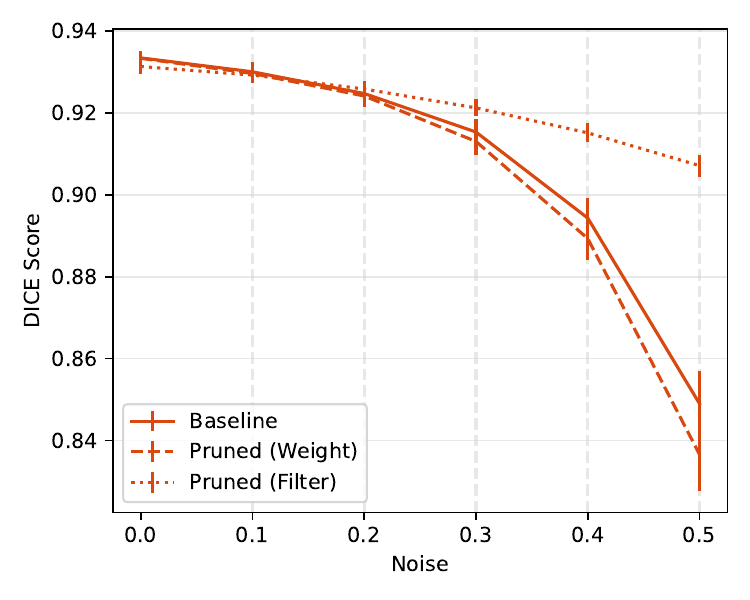}
    \end{minipage}\hfill
    \begin{minipage}{0.495\textwidth}
        \centering
        \includegraphics[width=\linewidth]{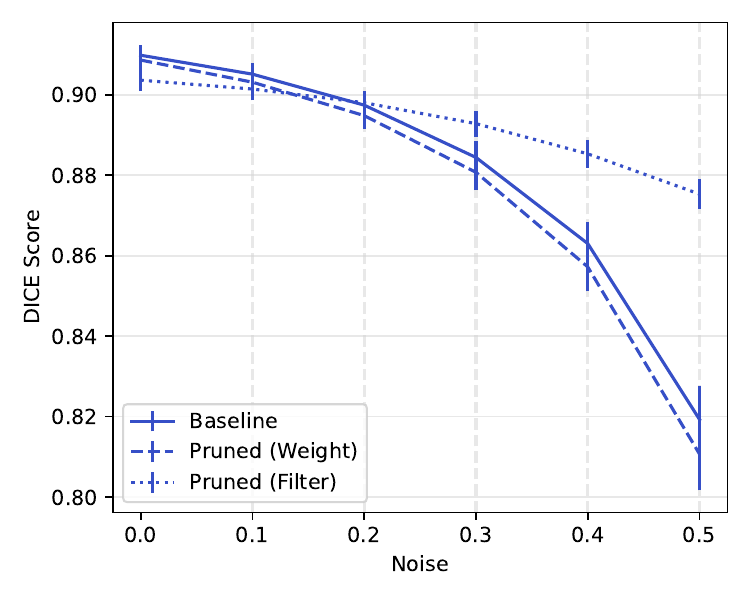}
    \end{minipage}
    \caption{Trend in robustness across both diastolic (left) and systolic (right) frames for the baseline, weight-pruned and filter-pruned models.}
    \label{fig:robustness}
\end{figure}

\textbf{Robustness} Figure \ref{fig:echo-noise-samples} details the degradation in the input image with an increasing noise ratio (i.e. the fraction of pixels removed). Echocardiograms are inherently noisy, both due to the scanner's quality and the sonographer's expertise; evaluating a model under increasing noise ratios is a simple step to help gauge its effectiveness under real-world scenarios. Figure \ref{fig:robustness} details the trends in robustness (quantified by the DICE score) for both the systolic and diastolic frames. Notably, we use the 87.5\% sparse weight-pruned model and the filter-pruned model with a filter fraction of 0.875. 

\begin{figure}[h]
    \centering
    \begin{minipage}{0.495\textwidth}
        \centering
        \includegraphics[width=\linewidth]{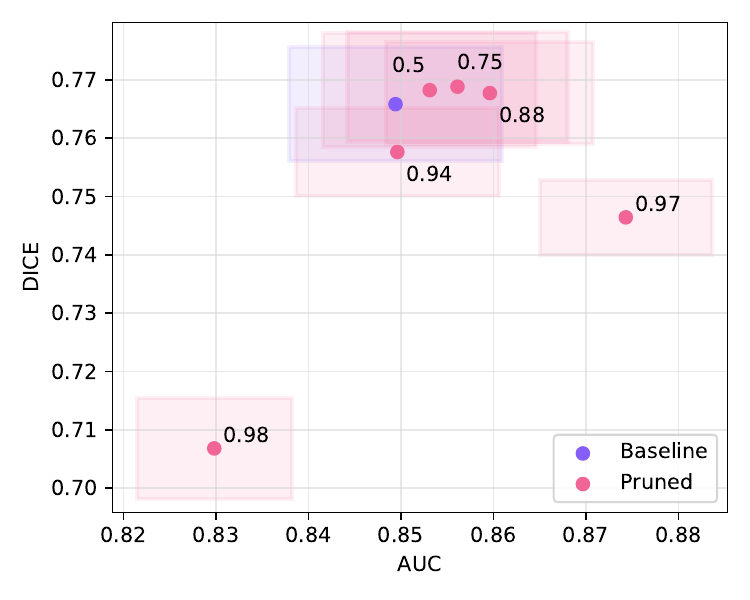}
    \end{minipage}\hfill
    \begin{minipage}{0.495\textwidth}
        \centering
        \includegraphics[width=\linewidth]{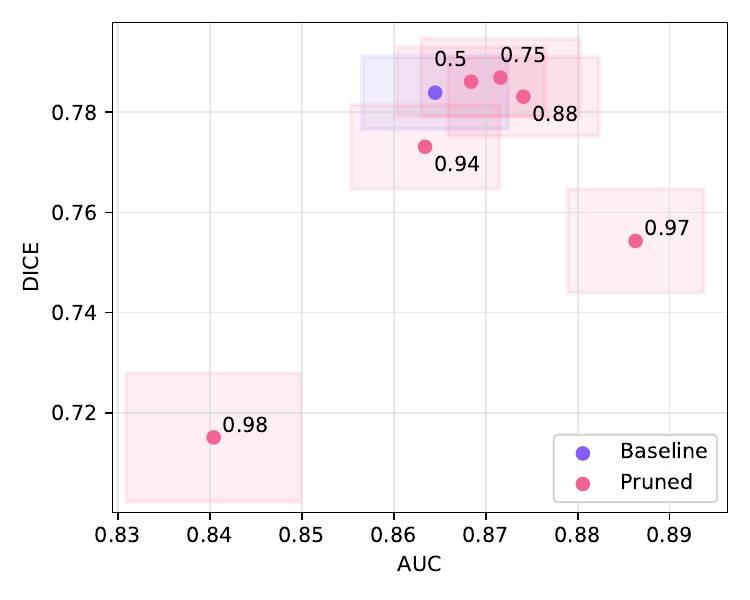}
    \end{minipage}
    \caption{Generalisability trends of models on an out-of-distribution dataset with two human annotations per sample. The translucent boxes represent error margins along each axis.}
    \label{fig:scatter}
\end{figure}

\textbf{Generalisability} Our test involved evaluating our retinal vessel segmentation models on an external dataset: DRIVE \cite{1282003}. Notably, for each sample, this dataset features two labels annotated by different human observers. Figure \ref{fig:scatter} showcases the trend in model quality across both labels through scatter plots; we use both DICE and AUC to identify the best model, generally located close to the top right. In our case, the model with a filter fraction of 0.875 generalises the best; the case with a fraction of 0.97 faces too much degradation in the DICE score. The model with a fraction of 0.75 is also a worthy contender, but notably, it is roughly 3x larger. While the actual scores are pretty low, we primarily examine the relative differences. Across both labels, the trend remains consistent.

\end{document}